\documentclass[letterpaper, 10 pt, conference]{ieeeconf}  

\IEEEoverridecommandlockouts                              

\usepackage{adjustbox}
\usepackage[font=small,belowskip=-1.5pt]{caption} 
\usepackage{tikz}
\usepackage[colorlinks=true]{hyperref}

\usepackage{amssymb}
\usepackage{mathtools} 
\usepackage{bbm}

\usepackage{pifont}
\usepackage{alphalph}
\usepackage{amsmath}
\usepackage{booktabs}
\usepackage{multirow}
\usepackage{interval}
\usepackage{caption}
\usepackage{subcaption}
\usepackage{makecell}

\definecolor{azure}{rgb}{0.0, 0.5, 1.0}

\newcommand{\ours}{\textsc{LOC-ZSON}}

\begin{document}

\title{\LARGE \bf

\ours{}: Language-driven Object-Centric Zero-Shot Object Retrieval and Navigation


}


\author{
Tianrui Guan$^{12*}$\thanks{\scriptsize * Work done during an internship at Amazon Lab126.}, 
Yurou Yang$^{1}$, 
Harry Cheng$^{1}$, 
Muyuan Lin$^{1}$, 
Richard Kim$^{1}$,\\ 
Rajasimman Madhivanan$^{1}$, 
Arnie Sen$^{1}$, 
Dinesh Manocha$^{2}$ 
\thanks{\scriptsize	$^{1}$Authors are with Amazon Lab126, Sunnyvale, CA 94089, USA. 
}%
\thanks{\scriptsize	$^{2}$Authors are with the Department of Computer Science, University of Maryland, College Park, MD 20742, USA. 
}
}

\maketitle
\thispagestyle{empty}
\pagestyle{empty}

\begin{abstract}
In this paper, we present \ours{}, a novel Language-driven Object-Centric image representation for object navigation task within complex scenes. 
We propose an object-centric image representation and corresponding losses for visual-language model (VLM) fine-tuning, which can handle complex object-level queries.
In addition, we design a novel LLM-based augmentation and prompt templates for stability during training and zero-shot inference. 
We implement our method on Astro robot and deploy it in both simulated and real-world environments for zero-shot object navigation. We show that our proposed method can achieve an improvement of 1.38 - 13.38\% in terms of text-to-image recall on different benchmark settings for the retrieval task. For object navigation, we show the benefit of our approach in simulation and real world, showing 5\% and 16.67\% improvement in terms of navigation success rate, respectively.

\end{abstract}


\section{Introduction}




In recent years, the concept of introducing assistant robots into household environments has transformed into a tangible reality, primarily due to the remarkable advancements achieved in the domains of computer vision, robotics, and human-computer interaction. One important capability for a home robot is to perform zero-shot object goal navigation (ZSON)~\cite{chaplot2020object}, the ability to locate and navigate to the object of interest in a scene based on a user inquiry, without training on this specific setting. The challenge of zero-shot object navigation (ZSON) has primarily been investigated from the perspective of image-goal navigation and exploration~\cite{2022zson, 2023zson, 2022cow}. From this perspective, their navigation policies are designed to autonomously generate actions and are trained through an end-to-end reinforcement learning paradigm. 
However, their focus on exploration for each episode limits the ability to retrieve objects based on past memory beyond that episode. 
In addition, it is difficult to analyze whether a failure case is caused by the image-text feature representation or the RL policy in those end-to-end navigation methods.

A critical aspect of ZSON involves aligning the representation of semantics (in form of text) with images, which is used for either specifying an image goal~\cite{2022zson} or use it as a stopping condition once the object is in current field of view during exploration~\cite{2022cow}. Given a target text, the pre-trained alignment model can assist ZSON methods in locating the desired objective within the image or high-dimensional semantic embedding space~\cite{2022zson}, even if the target object was not present during training. Those alignment models are often assumed due to the success in visual-language models (VLM) such as CLIP~\cite{clip}, and therefore is rarely explored in this context. However, the alignment model may occasionally retrieve or match with an undesired goal that does not correspond with the text query, resulting in a failed downstream navigation task. One possible reason for this failure could be that the compact image-level embeddings like CLIP~\cite{clip}, FLAVA~\cite{flava}, etc. may lose some object-level information. Alternatively, object grounding methods like OWL-ViT~\cite{owlvit} focus on objects and may resolve this issue through query-based detector, but this method is optimized for object retrieval over one single image, not for a database of images in object retrieval tasks.

\begin{figure}[t]
    \centering
    \includegraphics[width = \columnwidth]{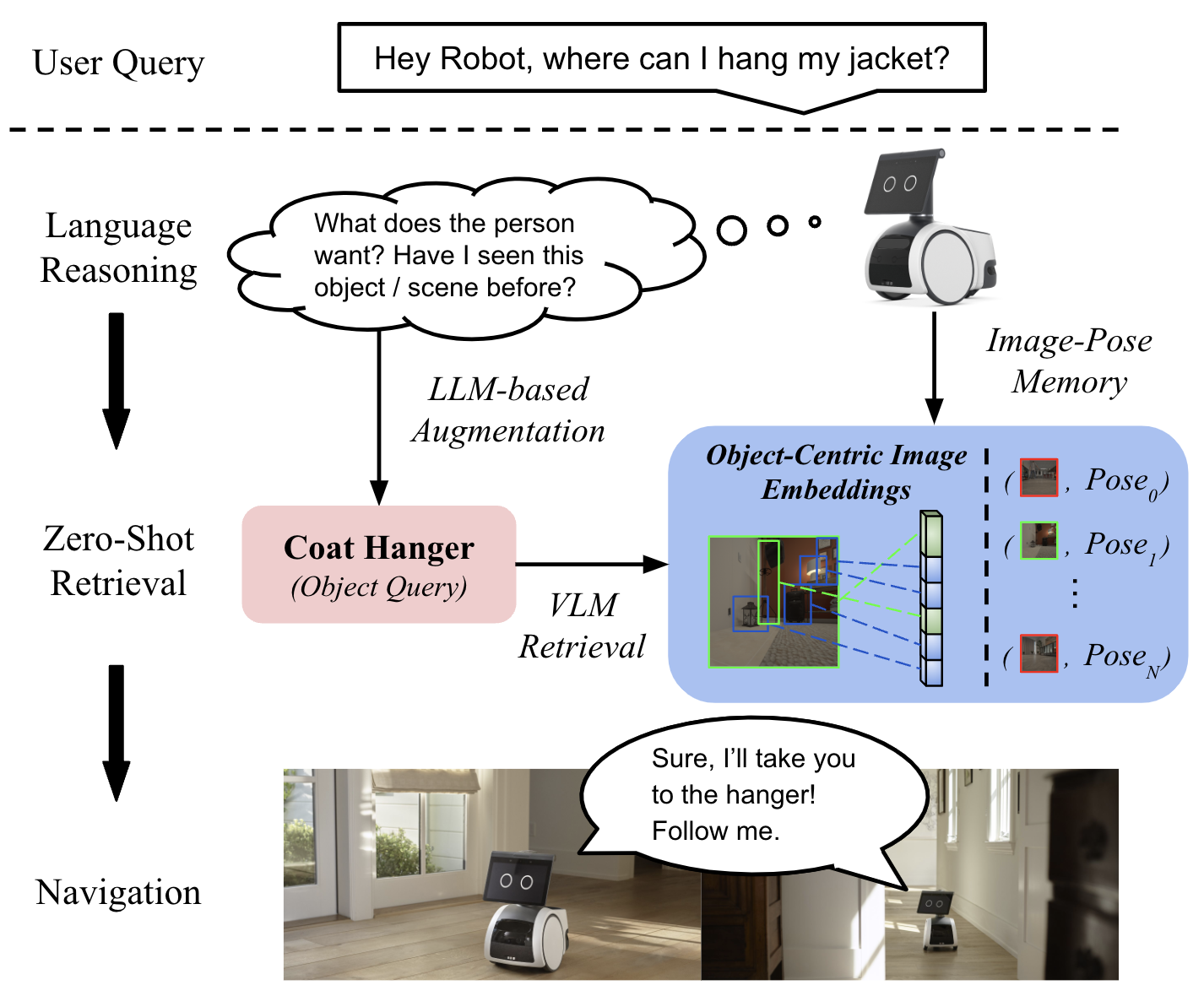}
    \caption{\textbf{Overview of the proposed \ours{}:} Our method performs Language-driven Zero-Shot Object Navigation (L-ZSON) in three steps: 1) Language Reasoning: we use LLM-based Augmentation and prompt engineering to parse the user query. 2) Zero-Shot Retrieval: we propose a novel object-centric image representation to localize the object from a database of images collected from exploration. 3) Navigation: The ground robot navigates to the top N candidates locations based on image-pose pairs.}
    \vspace{-7mm}
    \label{fig:cover}
\end{figure}

In this paper, we investigate the problem of performing the ZSON task and focus on the perspective of visual-language alignment in an indoor environment. 
To enhance accountability, rather than employing RL and end-to-end training, we decouple the object navigation task into two components, image-goal retrieval and navigation, placing particular emphasis on image-text alignment. 
To exploit past experience, our design leverages the benefits of both compact embedding representation in similarity-based alignment model like CLIP~\cite{clip}, and object-centric representation extraction in grounding methods like OWL-ViT~\cite{owlvit}.








\noindent\textbf{Main Results:} We introduce \ours{}, an innovative \textbf{L}anguage-driven \textbf{O}bject-\textbf{C}entric \textbf{Z}ero-\textbf{S}hot \textbf{O}bject \textbf{N}avigation paradigm, shown in Figure~\ref{fig:cover}. 
We approach the task of ZSON from a retrieval perspective and employ image-text retrieval with a novel object-level image representation. We improve the object navigation performance through accurately retrieving the corresponding objects from a memory database.
Our design and evaluation settings incorporate numerous features, including handling of a diverse range of natural language queries and zero-shot generalizations that do not require training in a specific indoor environment.

The key contributions of our work include:
\begin{enumerate}
    \item We propose a new object-centric image representation designed to capture object-level feature for image-text retrieval. We design an object-focused multi-label training loss to facilitate the convergence of this representation during fine-tuning. We outperform prior VLM models on text-to-image zero-shot object retrieval in terms of top $1$ average recall by $1.38 - 13.38\%$ in simulation and SUN RGB-D~\cite{sunrgbd} dataset.
    \item We introduce a data augmentation and prompting scheme grounded in Large Language Models (LLM) for fine-tuning Visual-Language Model (VLM). 
    Our innovative augmentation scheme effectively improves the performance stability during fine-tuning. In addition, we show that our prompting template consistently improves performance for all VLMs by at least $6.6 \%$ over a broad spectrum of natural language queries centered on objects during the inference phase.
    \item  We showcase the object navigation capabilities of our system in both simulated and real-world environments on a non-holonomic ground robot, and demonstrate improved object navigation success rate by 5\% and 16.67\% compared to other VLM-based methods~\cite{2022cow} in the simulation and real world, respectively.

\end{enumerate}

\section{Related Work}

\subsection{Zero-Shot Object Navigation (ZSON)}


%


ZER~\cite{al2022zero} uses a modular transfer learning model that first trains a semantic search policy for an image-goal task, then relates target goals to image goals via joint embedding training. 
CLIP-on-Wheels (CoW)~\cite{2022cow} uses a heuristic exploration policy and localizes objects using a gradient-based visualization technique with CLIP. 
Instead of explicitly localizing objects, \cite{2023zson} transform goal images into a multimodal, semantic embedding space and trains semantic-goal navigation agents. 
While the goal is to localize and approach the object of interest, ZSON~\cite{2022zson, 2023zson, 2022cow} has been defined under a reinforcement learning framework. Those methods usually start with exploration in an unknown environment, and only use experiences within each episode. In this paper, we discuss and address the task from an object retrieval perspective and focus on exploiting based on previous memory.

\subsection{Visual-Language Model (VLM)}
Vision language models bridge the gap between vision and language, and have shown capabilities on a wide array of applications, including visual question answering~\cite{antol2015vqa}, visual captioning~\cite{xie2022visual},
visual instruction tuning~\cite{visualinstruction}, and visual generation~\cite{feng2023layoutgpt}.
%
Despite hallucination issues~\cite{guan2023hallusionbench} and potential safety concerns~\cite{wu2024safety}, there is a growing utilization of vision language models in robotic applications, including manipulation~\cite{pgvlm2023} and navigation~\cite{llmnav}.
Through the incorporation of these models into robotic systems, the robot gains the capacity to directly interpret high-level language commands and improve its multimodal understanding, resulting in a more intuitive form of interaction between humans and robots~\cite{huang2023visual}.
Meanwhile, vision language models can potentially empower robots to operate in open-world scenarios~\cite{2022cow, 2022zson, clipbasednav}, due to knowledge transfer from large pre-trained model like CLIP~\cite{clip}. 
%


\subsection{Object-Centric Representation}
Object-centric representation focuses on identifying and modeling individual objects within a scene, allowing for more fine-grained understanding and reasoning about the elements present compared to scene-centric representations.
Previous studies in object-centric learning~\cite{burgess2019monet, engelcke2019genesis, greff2019multi, NEURIPS2021_43ec517d, lin2020space}  
have illustrated the feasibility of unsupervised training to decompose images into objects. 
MONet~\cite{burgess2019monet} employs a UNet architecture for mask extraction of individual objects, subsequently utilizing a Variational AutoEncoder (VAE) to generate reconstructed images of these objects.
%
%
SPACE~\cite{lin2020space} uses a foreground module to identify bounding boxes and reconstruct foreground objects and a background module to segment a scene into several components. 
%

\begin{figure*}[t]
    \centering
-    \includegraphics[width=\textwidth]{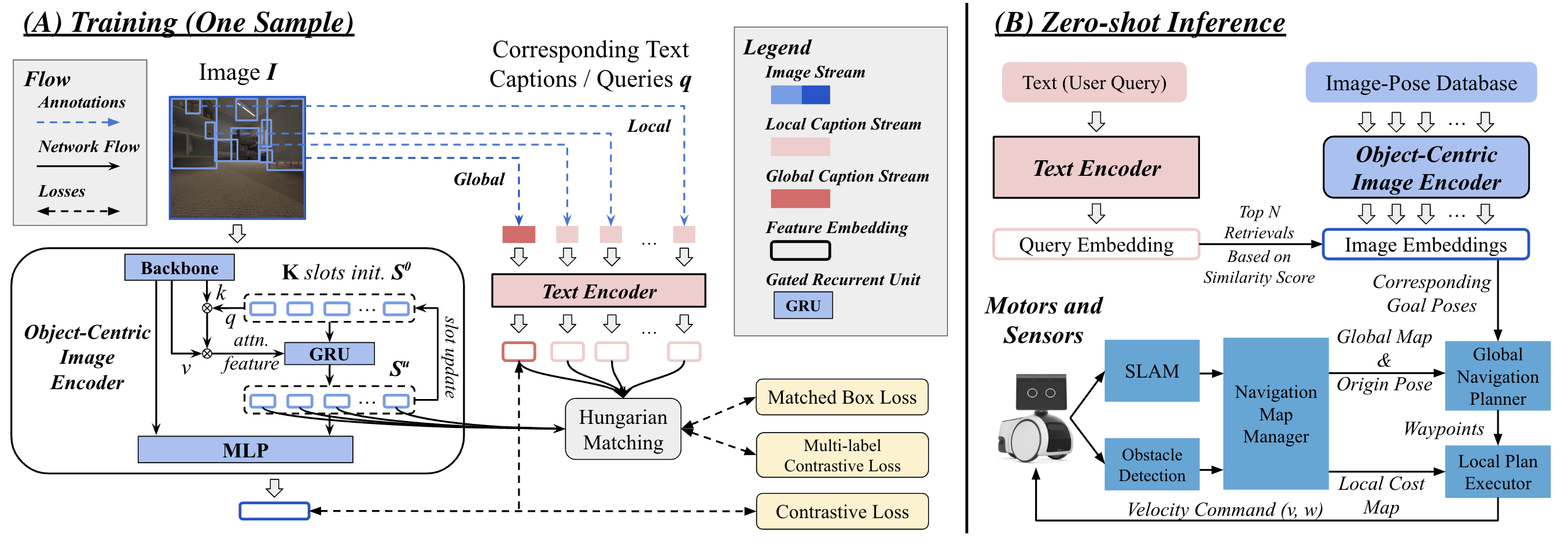}
    \caption{\textbf{Architecture of our proposed Object-Centric Image Encoder in \ours{}:} \textbf{(A)} We introduce a novel object-centric image representation with multi-label training. During training time, we feed one image with multiple object captions into our object-centric image encoder and text encoder. We use Hungarian matching algorithm~\cite{hungarian} to match local image feature with text annotation. \textbf{(B)} For inference, we send query with our prompt template into text encoder and retrieve the top image-pose pairs based on previous explorations. We pass the corresponding poses to the navigation stack for object navigation task. The image embeddings can be pre-computed. 
    }
    \vspace{-6mm}
    \label{fig:network}
\end{figure*}

\section{Method}

In this section, we explain our proposed \ours{} and \textbf{L}anguage-driven \textbf{O}bject-\textbf{C}entric image embedding. Next, we elaborate on the losses for fine-tuning and the prompting strategy. Finally, we describe our navigation components.

\subsection{Problem Definition}

In the context of Language-Driven Zero-Shot Object Navigation (L-ZSON), the goal is to navigate to designated objects identified through natural language description~\cite{2022cow}. The L-ZSON task has traditionally been framed within a Reinforcement Learning (RL) setting~\cite{2022zson, 2022cow} and uses pre-trained VLMs for feature extraction. To investigate and improve L-ZSON from the perspective of visual-text alignment, we introduce an alternative formulation with the same objective
as follows:


Let $(I_i, p_i)\in\mathcal{O}$ be a set of observations or memories in an indoor environment after some explorations, where $I_i$ is the RGB image and $p_i$ is the corresponding pose where the observation is made in the map frame $\mathcal{M}$. Let $\mathcal{Q}$ be a set of natural language sentences that explicitly or implicitly indicate an object or a general scene. Given the queries $\mathcal{Q}$, the goal is for the robot to navigate to a position such that the object of interest is within its field of view. 

In our proposed \ours{} approach, we decouple the task into retrieval and navigation stages. For image retrieval, given the query $\mathcal{Q}$ and an image database $\mathcal{I}$, the goal is to select the top $k$ relevant images $\{\bar{I}_0, ..., \bar{I}_{k-1}\}_i\subseteq\mathcal{I}$ for each $q_i\in\mathcal{Q}$. A true corresponding pair of $I_g\in\mathcal{I}$ and $q_g\in\mathcal{Q}$ indicates that the object described by $q_g$ is in the field of view of $I_g$. 
Once the object is localized in the image database $\mathcal{I}$, based on an image-pose pair from the memory, the robot can navigate to the desired location.

In contrast to image caption datasets~\cite{coco, flickr}, the images used for object navigation are collected sequentially and may feature multiple objects in a single view; therefore, it is possible for the same object to appear across multiple images. Conversely, it's also possible for a single image to contain several objects. We evaluate the zero-shot image-text retrieval capability of our method within this unique multi-label correspondence framework.



\subsection{Model Architecture for Object-Centric Retrieval}
\label{method:arch}

Our model architecture is shown in Figure~\ref{fig:network}. The model incorporates two key components: an object-centric image encoder for extracting visual representations and a text encoder for generating textual information.

\noindent\textbf{Object-Centric Image Encoder.} Given an image $I$, we resize and split it into fixed patches for the image encoder. We adopt ViT-B/16~\cite{vit} architecture for the image backbone and obtain $h^I_{last}\in\mathbb{R}^{D}$ from the output of the last layer.

To further extract object-level details, we propose an attention-based localization module, inspired by SlotAttention~\cite{slotatt}. Let $S^0 \in \mathbb{R}^{K\times D_{s}}$ be a list of $K$ initial slot embeddings initialized from a Gaussian distribution $\mathcal{N}(\mu, \text{diag}(\sigma))$ and $U$ be the number of iterations for the slot updates. During the $u$-th iteration, the slot features are updated as follows:
\vspace{-2mm}
\begin{equation}
    A^u = softmax(\frac{1}{\sqrt{D_s}}k(h^I_{last}) \cdot q(S^{u-1})^T) \in\mathbb{R}^{N\times K},
\end{equation}
\begin{equation}
W^u_{i, j} = \frac{A_{i, j}}{\sum^N_{l=1}A^u_{l, j}}\in\mathbb{R}^{N\times K},
\end{equation}
\begin{equation}
S^u = S^{u-1} + MLP(LN(G({W^u}^\mathbf{T}\cdot v(h^I_{last}), S^{u-1}))),
\end{equation}
where $S^u \in\mathbb{R}^{K\times D_{s}}$,  $LN$ is Layer Normalization, $q$, $k$, $v$ are linear projections with dimension $D_s$, and $G(f, H)$ is a Gated Recurrent Unit (GRU)~\cite{gru} with hidden dimension $D_s$ with $f$ as input features and $H$ as hidden states.

In the next step, we design a 2-stream class-agnostic detector that learns to focus on object localization and feature extraction from the corresponding region. After $U$ iterations, we use slot feature $S^U\in\mathbb{R}^{K\times D_{s}}$ to localize $K$ different sections of the images through a few MLP layers for bounding box regression and obtain $\mathbf{B_I}\subseteq\mathbb{R}^{K\times 4}$. We consider $S^U$ to contain local feature information, as it can be decoded into different regions with a detection head. Finally, we obtain image embedding $e^I$ by aggregating image-level embedding $h^I_{last}$ and object-level embedding $S^U$ as follows:
\vspace{-2mm}
\begin{equation}
e^I = MLP(concate(h^I_{last}, linear(S^U))) \in\mathbb{R}^{D},
\end{equation}
Our object-centric image encoder tries to solve one key challenge for image-text retrieval task with multiple object labels, which is that the image representation should contain object-level information and produce high similarity scores over different object queries simultaneously. In most existing VLM frameworks~\cite{clip, flava, align}, the image embedding is trained and optimized over a single corresponding caption at a time, which could not apply to our case; otherwise, the model may not converge after trained on the same image with multiple different captions describing separate objects.

\noindent\textbf{Text Encoder.} Given a text query $q$, we tokenize and convert it into word vectors according to BERT~\cite{bert}. Similar to our visual backbone, we employ the ViT-B/16 architecture to encode these word vectors. Then we apply a linear projection on the output of the last transformer layer $h^T_{last}$ and use it as the text embedding $e^T\in\mathbb{R}^{D}$ for the retrieval task.


\noindent\textbf{Zero-Shot Image-Text Retrieval.} 
From $N$ images and $M$ text queries, we can obtain a list of image embeddings $E^I\in\mathbb{R}^{N\times D}$ and text embeddings $E^T\in\mathbb{R}^{M\times D}$. We can compute a similarity score $Sim = E^T {E^I}^\mathbf{T}$. For text-to-image retrieval, the indexes of the top $k$ images that are most relevant to the $i$-th text query are:
\vspace{-1mm}
\begin{equation}
index^I_i = argtopk(Sim[i, :], k) ,
\end{equation}
Similarly, for image-to-text retrieval, the indexes of top $k$ queries that are most relevant to the $j$-th image is:
\vspace{-1mm}
\begin{equation}
index^T_j = argtopk(Sim[:, j], k) ,
\end{equation}



\subsection{Visual-Language Fine-tuning with Multi-Correspondence}

Based on the architecture described in Section~\ref{method:arch}, we propose a novel training approach with several objectives that deal with the multi-labels problem, \textit{i.e.}, a single image with multiple text correspondences for each object in the image. Let the training annotation $(q_g, b_g)\in\mathbf{G_I}$ be the text-bounding box pair in image $I\in\mathcal{I}$.
Our proposed training approach involves the following components and losses. 

\noindent\textbf{Contrastive Loss.} Similar to CLIP~\cite{clip}, given a batch of images and text, we maximize the cosine similarities between matched image and text pairs and minimize those for the unmatched pairs. To convert the multiple labels into one label in our setting, we simply randomly permute the text labels and concatenate them before passing into the text encoder. The contrastive loss 
$\mathcal{L}_{C}$ of encoder 
outputs $e^I$ and $e^{T_{cat}}$ is calculated as: 
\vspace{-2mm}
\begin{equation}
\mathcal{L}_{C}(e^I, e^{T_{cat}}, y) = CrossEntropy(e^I\cdot e^{T_{cat}}, y),
\end{equation}
where $y=1$ if $I$ and ${T_{cat}}$ are matched pairs, and $0$ otherwise.


\noindent\textbf{Hungarian Matching.} Given predicted boxes $\mathbf{B_I}$ of size $K$ and ground truth $\mathbf{G_I}$ of size $N$, we want to find the optimal assignment $\mathcal{A}$:
\vspace{-4mm}
\begin{align}
\text{Minimize} \quad & \sum_{i=1}^{K} \sum_{j=1}^{N} d(\mathbf{B_I}[i], \mathbf{G_I}[j]) \cdot x_{ij} \\
\text{Subject to} \quad & \sum_{j=1}^{N} x_{ij} \leq 1, \quad \forall i = 1, \ldots, K \\
& \sum_{i=1}^{K} x_{ij} \leq 1, \quad \forall j = 1, \ldots, N \\
& x_{ij} \in \{0, 1\},
\end{align}
where $d= \mathcal{L}_{L1} + \mathcal{L}_{GIoU}$, explained in \textit{Matched Box Loss}.

The Hungarian algorithm~\cite{hungarian} resolves the problem of optimal assignment and can be applied to the problem of matching predicted bounding boxes with ground truth bounding boxes in detection tasks. 
Once we find the optimal assignment $\mathcal{A}$ using hungarian algorithm, we can add some local losses to improve the object-level representation of the image embedding.

\noindent\textbf{Matched Box Loss.} Based on the optimal assignment $\mathcal{A}$, we use two losses for the matched bounding boxes: 
regression loss $\mathcal{L}_{L1}$ and Generalized Intersection over Union (GIoU) Loss $\mathcal{L}_{GIoU}$. 
\vspace{-3mm}
\begin{equation}
    \mathcal{L}_{L1}(x_1, x_2) = \sum_{i=0}^{3} |x_1[i] - x_2[i]|,
\end{equation} \vspace{-2mm}
\begin{equation}
    \mathcal{L}_{GIoU}(x_1, x_2) = \frac{A_{\text{intersection}}}{A_{\text{union}}} - \frac{A_{\text{enclosing}} - A_{\text{union}}}{A_{\text{enclosing}}},
\end{equation}
where $A_{\text{intersection}}, A_{\text{union}}, A_{\text{enclosing}}$ are the area of the intersection, union, and smallest convex hall of the two boxes.

\noindent\textbf{Multi-label Contrastive Loss.} Since each slot feature in $S^U$ corresponds to one box prediction, based on $\mathcal{A}$, we can find the correspondence between slot features and the corresponding ground truth text labels. Let $\{e\}^T_I$ be a list of text embeddings after the ground truth text labels of image $I$ are fed to the text encoder, $S^U_I$ be the slot features of image $I$, and $\mathcal{A}_I$ be the assignment from predicted slots to the ground truth text labels. The multi-label contrastive loss $\mathcal{L}_{MC}$ is calculated as:
\vspace{-1mm}
\begin{equation}
\mathcal{L}_{MC}(S^U_I, \{e\}^T_I, \mathcal{A}_I) = CrossEntropy(\mathcal{A}_I(S^U_I), \{e\}^T_I).
\end{equation}
The final loss $\mathcal{L} = \alpha \mathcal{L}_{C}+ \beta \mathcal{L}_{L1}+ \gamma \mathcal{L}_{GIoU}+ \delta \mathcal{L}_{MC}$, where $\alpha, \beta, \gamma, \delta$ are weighting coefficients.

\subsection{VLM Prompting and LLM-based Data Augmentation}

In this section, we discuss one important aspect that our object navigation pipeline aims to handle, which is natural language understanding through prompting. 
In practical scenarios, natural language and queries exhibit a wide range of diversity, and the expressions used can sometimes be implicit or nuanced.
During inference, a query text could be a description of an object, a question asking for the object, or simply a noun. However, one small change on the text query for the VLM could significantly affect the retrieval performance. For example, we find that we can improve the retrieval success rate on a specific object by simply adding/removing definite and indefinite articles or changing its singular form to plural form (or vice versa). 
To solve this issue, we propose an LLM-based generation method for  data augmentations, and design a prompting template with more context for more stable performance of the VLM.

\noindent\textbf{Noun-to-Sentence Generation.} Given an object noun, we ask the GPT-3.5 to generate sentences or questions that describe or ask for this item. To generate a diverse set of sentences, all previous generations are added to the GPT-3.5 input to ensure the output queries are different from previous generations. We also specify that the sentences should explicitly or implicitly refer to this object noun to improve the diversity of the data.

\noindent\textbf{Sentence-to-Noun Generation.} Similarly, given the sentence or description, we ask GPT-3.5 to produce a short noun as an answer so that the object goal is explicitly extracted or inferred based on the text reasoning of GPT. This step is extremely important and useful when the sentence is implicitly referring to an object.

\noindent\textbf{Multi-labels Data Augmentation.} Most existing datasets with image-text pairs like COCO~\cite{coco} and Flickr30K~\cite{flickr} are general image caption datasets, and there are limited indoor datasets that have object-level caption annotations. With our LLM-based generation pipelines, we convert indoor detection datasets like SUNRGBD~\cite{sunrgbd} into multi-label image-caption pairs, and we can easily scale up the number of captions on each object in the image for training and evaluation.

\noindent\textbf{VLM Prompting.} Based on the above two generation pipelines, we design the prompt template for the VLM as: \textit{the object noun} followed by \textit{the query text and sentences}. There should be many ways to improve and stabilize the performance of VLM further, but this design seems to be the optimal among a few simple prompting strategies. 
In addition to inference, we integrate this prompt design for VLM fine-tuning, and it also leads to significant improvement, as demonstrated in Section~\ref{sec:ablation}.

\subsection{Language-Driven Zero-Shot Object Navigation (L-ZSON)}
The navigation stack uses both depth and stereo cameras, and contains 
three components: \textit{mapping pipeline}, \textit{global navigation planner}, and \textit{local plan executor}. 
The \textit{mapping pipeline} integrates state-of-the-art SLAM system~\cite{slam_survey} for mapping and localization, and it would maintain and update a layered global occupancy map propagated with obstacles and objects captured by depth cameras for collision avoidance. The \textit{navigation map manager} interprets the global occupancy map as cost and feeds it into the path searching algorithms.
When a ``go to" command is given, the \textit{global navigation planner} would  generate a global plan towards the target point and the \textit{local plan executor} would proceed along the provided waypoints. With these components, the robot can execute commands like ``explore and mapping," ``go to", and ``find" in an indoor environment.

Based on the image-pose pairs $(I, p)\in\mathcal{O}$ from exploration, we use the ``go to" command to navigate to the top $k$ 
poses and stop when we localize the object in view. 
\section{Implementation Details and Analysis}

\begin{table*}[t]
\centering
\Large
\begin{adjustbox}{max width=\textwidth}
\begin{tabular}{c|c|c|cc|cc|cc|cc|cc}
  \toprule[2pt]
  
  \multirow{3}[2]{*}{\textbf{\rotatebox[origin=c]{0}{\makecell{Retrieval \\ Task}}}} & \multirow{3}[2]{*}{\textbf{Method}} & \multirow{2}[1]{*}{\textbf{\rotatebox[origin=c]{0}{\makecell{Num. \\ of \\ Param.}}}} & \multicolumn{2}{|c|}{\multirow{2}[1]{*}{\textbf{Simulation}}} &  \multicolumn{8}{c}{\textbf{SUN RGB-D~\cite{sunrgbd}}} \\
        \cline{6-13}
    & & & & & \multicolumn{2}{c|}{\textbf{kv1}} & \multicolumn{2}{c|}{\textbf{kv2}} & \multicolumn{2}{c|}{\textbf{realsense}} & \multicolumn{2}{c}{\textbf{xtion}}\\
    \cline{4-13}
    & & & \textbf{AR@1} $\uparrow$  & \textbf{AR@5} $\uparrow$ & \textbf{AR@1} $\uparrow$  & \textbf{AR@5} $\uparrow$ & \textbf{AR@1} $\uparrow$  & \textbf{AR@5} $\uparrow$ & \textbf{AR@1} $\uparrow$  & \textbf{AR@5} $\uparrow$ & \textbf{AR@1} $\uparrow$  & \textbf{AR@5} $\uparrow$\\
\hline

    \multirow{9}[5]{*}{\rotatebox[origin=c]{0}{\makecell{Text \\ To \\ Image}}} & CLIP-base~\cite{clip} & 151 M & 42.86 & 54.89 & 15.94 & 29.82 & 11.80 & 23.34 & 20.6 & 41.28 & 12.45 & 25.09 \\
    & CLIP-large~\cite{clip} & 428 M & 43.36 & 58.40 & 18.65 & \underline{33.81} & \underline{13.95} & \underline{26.94} & \underline{24.85} & \underline{44.77} & \underline{14.35} & 27.87 \\
    & BLIP-image-caption~\cite{blip} & 225 M & 15.29 & 50.38 & 0.91 & 3.23 & 1.61 & 5.23 & 2,72 & 6.98 & 0.73 & 3.26 \\
    & BLIP-retrieve-large~\cite{blip} & 447 M & 31.58 & \textbf{82.21} & 1.6 & 4.32 & 0.31 & 2.83 & 2.64 & 9.7 & 0.59 & 2.05 \\
    & ALIGN~\cite{align} & 172 M & \underline{53.88} & 58.4 & \textbf{19.02} & 22.26 & 10.75 & 24.65 & 23.23 & 42.13 & 13.33 & \underline{28.29} \\
    & FLAVA~\cite{flava} & 241 M & 47.12 & 63.41 & 12.48 & 28.04 & 10.63 & 21.8 & 19.62 & 37.8 & 11.71 & 24.64 \\
    & OWL-ViT-base~\cite{owlvit} & 153 M & 12.28 & 13.78 & 3.23 & 5.52 & 3.08 & 5.32 & 4.26 & 6.81 & 2.97 & 6.36 \\
    & OWL-ViT-large~\cite{owlvit} & 434 M & 15.79 & 23.81 & 3.89 & 7.52 & 2.83 & 5.89 & 6.47 & 11.23 & 3.2 & 6.68 \\


    & \textbf{\ours{} (Ours)} & 377 M & \textbf{60.65} & \underline{74.94} & \underline{18.81} & \textbf{36.84} & \textbf{18.03} & \textbf{33.29} & \textbf{38.23} & \textbf{60.92} & \textbf{15.73} & \textbf{30.92} \\

 \midrule

    \multirow{9}[5]{*}{\rotatebox[origin=c]{0}{\makecell{Image \\ To \\ Text}}} & CLIP-base~\cite{clip} & 151 M & 42.86 & 54.89 & 23.87 & 49.14 & 21.13 & 49.14 & 30.59 & 67.48 & 22.1 & 46.03 \\
    & CLIP-large~\cite{clip} & 428 M & 45.61 & 79.6 & 21.61 & 45.27 & 17.96 & 47.04 & 18.36 & 56.12 & 24.82 & 46.39 \\
    & BLIP-image-caption~\cite{blip} & 225 M & 20.75 & 72.01 & 0.43 & 2.8 & 0.7 & 4.89 & 1.92 & 10.14 & 0.12 & 1.54 \\
    & BLIP-retrieve-large~\cite{blip} & 447 M & 34.05 & 65.81 & 1.4 & 20.32 & 2.26 & 16.67 & 1.22 & 15.38 & 0.89 & 11.32 \\
    & ALIGN~\cite{align} & 172 M & 50 & 74.72 & \textbf{49.57} & \textbf{77.42} & 31.02 & 62.74 & 37.06 & 67.83 & \textbf{32.7} & \underline{60.96} \\
    & FLAVA~\cite{flava} & 241 M & 53.55 & 77.44 & 27.85 & 57.96 & 23.17 & 55 & 40.56 & 74.16 & 27.07 & 53.08 \\
    & OWL-ViT-base~\cite{owlvit} & 153 M & 35.86 & 83.22 & 35.59 & 68.06 & \textbf{51.61} & 78.71 & \textbf{66.61} & 85.31 & 26.78 & 58.23 \\
    & OWL-ViT-large~\cite{owlvit} & 434 M & \textbf{67.2} & \textbf{98.61} & 29.57 & \underline{74.3} & 43.17 & \underline{78.98} & 54.37 & \underline{86.19} & 25.77 & \textbf{66.35} \\


    & \textbf{\ours{} (Ours)} & 377 M & \underline{61.84} & \underline{87.26} & \underline{37.2} & 70.43 & \underline{45.43} & \textbf{79.09} & \underline{60.83} & \textbf{88.99} & \underline{28.2} & 55.51 \\

  \bottomrule[2 pt]
\end{tabular}

\end{adjustbox}
\caption{\textbf{SOTA comparisons in simulation and the SUN RGB-D~\cite{sunrgbd} dataset:} We show the performance of \ours{} and SOTA VLMs in text-to-image and image-to-text retrieval tasks. We \textbf{bold} the best score and \underline{\textit{underscore}} the second best. We show that our method achieves the best or second best average recalls in most cases among SOTA methods, including ones with larger number of parameters. All results are obtained using their respective best prompts, among the prompting formats mentioned in Table~\ref{tab:aug_ablation}.
}
\vspace{-6mm}
\label{tab:comp}
\end{table*}

  




\subsection{Data Preparations and Evaluation Benchmark}

We use the following datasets for training and evaluation:

\noindent\textbf{SUN RGB-D~\cite{sunrgbd}.} The original dataset focuses on indoor scenes with multiple objects and detection annotations. We modified the dataset with our LLM-based text generation pipeline so each object label is augmented with multiple natural language queries and context to simulate real-world natural query scenarios.  The training and testing sets contain 5285 and 5050 images, respectively. There are a total of 1067 object text labels, and we scale up the number of different text contexts into 10670 queries for training and evaluation.

Note that this dataset can be used as a good training dataset for text association, but  there are some limitations for evaluation purpose; some of the object text labels refer to the same object but with different labels. As a result, the ground truth images only count the corresponding labels as correct without considering other labels with the same meaning.

\noindent\textbf{Simulation.} We collect 5000 images with a ground robot in our Unreal simulated indoor environment with segmentation ground truth and sample 1436 images as image database. We select 20 objects appearing in this environment and generate 400 different natural language queries for image-text retrieval. We obtain the ground truth pairs based on instance segmentation annotations and carefully screen each images, making sure the annotation fir each correspondence is valid. The list of objects includes furniture, amenities, human, decorations and other items.
We use this setting for zero-shot retrieval and object navigation evaluation after fine-tuning our method on the entire SUN RGB-D dataset.

\noindent\textbf{Real-world.} We collected 668 images in the real world. Since there is no ground truth annotation, we use it for qualitative visualization and zero-shot object navigation testing.

\noindent\textbf{Evaluation Metrics.} We evaluate the performance in terms of image retrieval and object navigation. 
\begin{itemize}
    \item Top $k$ Average Recall: Calculated as the number of successful top $k$ retrievals divided by the total number of test queries. Given a query, a top $k$ retrieval is successful if one of the top $k$ candidates is within the ground truth pairs.
    We measure the recall for both text-to-image and image-to-text retrieval tasks.
    \item Success Rate (SR): 
    We compute the SR by assessing two criteria: whether the robot successfully stops within $1$ meter or $2$ meters of the targeted object, and whether that object falls within the robot's field of view.
\end{itemize}

\subsection{Implementation Details}

Our implementation is built on the FLAVA~\cite{flava} and uses pre-trained FLAVA model to initialize. The models are fine-tuned on 8 NVIDIA RTX A5000 with a batch size of 32. We use the Adam optimizer with a learning rate of \(1e{-5}\) and exponential decay of \(1e{-2}\). We train our model for 50k iterations and use 1000 iterations to warm up. We freeze the text encoder during training. We interpolate the size of the images into $224\times 224$ and get $16\times 16$ patches for the ViT-based image backbone. We use $D = D_s = 768$ as the feature dimension of the embeddings. For slot attention, we set $\mu = \vec{0}, \sigma = \vec{1}$ for slots initialization and set the number of slots $K = 10$ and the number of iterations $U = 20$. For the loss, we set $\alpha = \beta = \gamma = \delta = 1$. All ablation studies are conducted on simulation data.

\subsection{Comparisons and Ablation Study}
\label{sec:ablation}
\noindent\textbf{State-of-the-art Comparisons.} In Table~\ref{tab:comp}, we show the results of state-of-the-art VLMs and our method in 5 different settings from our simulation and SUN RGB-D~\cite{sunrgbd}. Our method outperforms other SOTA methods by $1.38 - 13.38\%$ in most of the settings in terms of top 1 average text-to-image average recall. OWL-ViT~\cite{owlvit} is an object grounding method; therefore, we iterate over all images conditioned on the query, and choose the images that contain box with highest confidence. Note that OWL-ViT has very good image-to-text recall due to its detection capability, while has limited ability to retrieval images based on text. All results are performed in zero-shot settings, and testing data is not used for training. 

\noindent\textbf{Ablation on multi-labels, slot attention.} In Tables~\ref{tab:slot_ablation}, we show the effect of our components. 
We observe that when used individually, each component results in performance degradation; when combined, they collectively yield a improvement of $13.53 \%$.
One explanation is that slot attention is not fully effective without local supervision on each object.



\begin{table}[t]
\centering
\resizebox{\columnwidth}{!}{%
\begin{tabular}{cccccc}
  \toprule
  \multirow{2}[1]{*}{\textbf{\rotatebox[origin=c]{0}{\makecell{Init. with \\ Pre-trained}}}} & \multirow{2}[1]{*}{\textbf{\rotatebox[origin=c]{0}{\makecell{Fine-tuning \\ Iterations}}}} & \multirow{2}[1]{*}{\textbf{\rotatebox[origin=c]{0}{\makecell{Multi \\ Labels}}}} & \multirow{2}[1]{*}{\textbf{\rotatebox[origin=c]{0}{\makecell{Slot \\ Attn.}}}}   & \multicolumn{2}{c}{\textbf{Text-to-Image}}\\
  \cmidrule{5-6}
    &  &  &    & \textbf{AR@1} $\uparrow$  & \textbf{AR@5} $\uparrow$\\
 \midrule
     \checkmark & \ding{55}  & -  & - & 47.12 & 63.41  \\   
     \checkmark & 20K  & \ding{55}  & \ding{55} & 50.88 & 62.91  \\   
     \checkmark & 20K  & \ding{55}  & \checkmark  & 49.87 & 63.41 \\   
     \checkmark & 20K  & \checkmark  & \ding{55}  & 47.87 & 64.41 \\   
     \checkmark & 20K  & \checkmark  & \checkmark & \textbf{60.65} & \textbf{74.94} \\   
  \bottomrule
\end{tabular}
}
\caption{\textbf{Ablation study on model design:} We show the effect of multi-label training and slot attention in our object-centric design. 
}

\label{tab:slot_ablation}

\vspace{-3.7 mm}
\end{table}

\noindent\textbf{Ablation on losses.} In Tables~\ref{tab:loss_ablation} ,we observe that with only matched box loss or multi-label contrastive loss lead to performance drop. One possible explanation is that, training with only contrastive loss minimize the similarity between matched pairs, directly optimizing the retrieval objective; 
although both matched box loss and multi-label contrastive loss concentrate on local regions, they fall short in their respective capabilities: the former struggles to effectively extract local features, while the latter has limitations in identifying localized corresponding regions.

\begin{table}[t]
\centering
\resizebox{\columnwidth}{!}{%
\begin{tabular}{cccccc}
  \toprule
  \multirow{2}[2]{*}{\textbf{\rotatebox[origin=c]{0}{\makecell{Contrastive \\ Loss  }}}} & \multirow{2}[2]{*}{\textbf{\rotatebox[origin=c]{0}{\makecell{Box L1 \\ Loss  }}}} & \multirow{2}[2]{*}{\textbf{\rotatebox[origin=c]{0}{\makecell{Box GIoU \\ Loss  }}}} & \multirow{2}[2]{*}{\textbf{\rotatebox[origin=c]{0}{\makecell{Multi-label \\ Contrastive Loss  }}}}   & \multicolumn{2}{c}{\textbf{Text-to-Image}}\\
  \cmidrule{5-6}
    &  &  &    & \textbf{AR@1} $\uparrow$  & \textbf{AR@5} $\uparrow$\\

 \midrule
     \checkmark & \ding{55}  & \ding{55}  & \ding{55}  & 46.62 & 62.91 \\   
     \checkmark & \ding{55}  & \ding{55}  & \checkmark  & 45.36 & 64.91 \\   
     \checkmark & \checkmark  & \checkmark  & \ding{55}  & 44.61 & 56.39 \\   
     \checkmark & \checkmark  & \checkmark  & \checkmark  & \textbf{60.65} & \textbf{74.94} \\   
  \bottomrule
\end{tabular}
}
\caption{\textbf{Ablation study on losses:} We show the effects of each loss, which lead to $14.03 \%$ improvment in terms of AR@1.
}

\label{tab:loss_ablation}

\vspace{-6.5mm}
\end{table}

\noindent\textbf{Ablation on augmentation and prompting.} From each row of Table~\ref{tab:aug_ablation}, we can see that using the prompt template \textit{object noun + query sentence} consistently improves on all methods. In the last column, our augmentation strategy based on this prompting templates leads to $6.77 \%$ in terms of text-to-image AR@1.
Moreover, we discover that fine-tuning the model using complete sentences (\textit{v.s.} noun only) enhances the model's convergence compared to the pre-trained FLAVA baseline. This improvement likely occurs because the text encoder is originally pre-trained on full sentences.

\begin{table}[t]
\centering
\resizebox{\columnwidth}{!}{%
\begin{tabular}{ccccc}
  \toprule
  \multirow{2}[1]{*}{\textbf{\rotatebox[origin=c]{0}{\makecell{Text-to-Image \\ {AR@1}}
}}} & \multirow{2}[1]{*}{\textbf{\rotatebox[origin=c]{0}{\makecell{Training \\ Samples}}}} & \multicolumn{3}{c}{\textbf{Prompting Template for Zero-Shot Retrieval}}\\
  \cline{3-5} \\[-1.5ex]

   & & {\textbf{O}bject \textbf{N}oun} & {\textbf{Q}uery \textbf{S}entence}   & \textit{\textbf{ON} + \textbf{QS}} \\
    \cline{1-5} \\[-1.5ex]

     CLIP~\cite{clip} & -  & \underline{36.13} & 39.35 & 43.36 \\   
     ALIGN~\cite{align} & -  & 35.48 & \underline{50.13} & 53.88 \\   
     FLAVA~\cite{flava} & -  & \textbf{40.52} & 47.37 & 47.12 \\   
     OWL-ViT-Large~\cite{owlvit} & -  & 12.26 & 15.79 & 15.79 \\   
   \hline \\[-1.5ex]
     \multirow{4}[1]{*}{\textbf{\rotatebox[origin=c]{0}{\makecell{\ours{} \\ (Ours)}}}} & \textit{\textbf{O}bject \textbf{N}oun}  & 36.6 & 44.86 & 47.87\\   
     & \textit{\textbf{Q}uery \textbf{S}entence}  & 28.76 & 39.85 & 48.12\\   
     & \textit{\textbf{QS} w/ Aug.}  & 24.84 & 49.87 & \underline{54.89} \\   
     & \textit{\textbf{ON} + \textbf{QS} w/ Aug.}  & 34.64 & \textbf{55.89} & \textbf{60.65} \\  


  \bottomrule
\end{tabular}
}
\caption{\textbf{Ablation study on augmentation and prompting:} We show the effect of data augmentation and prompting template. 
}

\label{tab:aug_ablation}

\vspace{-4mm}
\end{table}


\label{tab:aug_ablation}



\subsection{Object Navigation with Non-Holonomic Ground Robot}

\begin{table}[t]
\centering
\resizebox{\columnwidth}{!}{%
\begin{tabular}{ccccc}
  \toprule




     \multirow{2}[1]{*}{\textbf{\rotatebox[origin=c]{0}{Method}}} & \multicolumn{2}{c}{\textbf{Success Rate in Simulation}} & \multicolumn{2}{c}{\textbf{Success Rate in Real World}} \\
\cline{2-5} \\[-1.5ex]
    & \textbf{Goal\textit{ (1m / 2m)}} & \textbf{Obj. in FOV} &  \textbf{Goal\textit{ (1m / 2m)}} & \textbf{Obj. in FOV} \\

   \hline \\[-1.5ex]
     CoW (CLIP)~\cite{2022cow}   & 10 / 27.5 & 35 & 25 / 58.33 & 75 \\  
     CoW (OWLViT)~\cite{2022cow}   & 2.5 / 20 & 30 & 25 / 50 & 58.33 \\   
   \hline \\[-1.5ex]
     \textbf{\ours{} (Ours)}   & \textbf{15} / \textbf{40} & \textbf{67.5} & \textbf{41.67} / \textbf{66.67} & \textbf{83.33} \\   
  \bottomrule
\end{tabular}
}
\caption{\textbf{Object navigation comparisons:} We compare \ours{} with two VLM-methods~\cite{clip, owlvit} in CoW~\cite{2022cow}.
For each method, we run 40 trials in the simulated environment and 12 trails in the real world on different objects based on top 1 prediction.
}





\label{tab:nav_comp}

\vspace{-6mm}
\end{table}

We show some navigation results in a simulated environment and the real world. In Table~\ref{tab:nav_comp}, we show that our method can outperform other VLM-based methods~\cite{2022cow} for the task of object navigation. Please refer to the video in the supplemental for qualitative results.


\section{Conclusions, Limitations, and Future Works}


In this paper, we present \ours{}, a zero-shot object navigation paradigm that uses a novel Language-driven Object-Centric image representation to capture object-level details, which improves the image retrieval and leads to better object navigation success rates. We present an LLM-based data augmentation and prompting pipeline that resolves the issue of stability during VLM fine-tuning. We demonstrate improved retrieval and navigation results in public benchmark dataset, a simulated and real home environment. 

Currently, our image representation is only used for retrieval tasks. In the future, we plan to explore its usage in an end-to-end RL framework like in~\cite{2023zson} and see whether our object-centric image representation can be beneficial to exploitation, and explore more prompting strategies~\cite{wang2023plar} to fully harness the power of VLM.



{\small
\bibliographystyle{IEEEtran}
\bibliography{citation}
}

\end{document}